# Maximal Margin Distribution Support Vector Regression with coupled Constraints-based Convex Optimization

**Gaoyang Li [1], Jinyu Yang [2], Chunguo Wu[3,*], and Qin Ma[4,*]**

[1] Shanghai Tenth People's Hospital, Tongji University, Shanghai, 200092 , China.[2] Computer Science And Engineer, University of Texas at Arlington, Arlington, 76019, USA.[3] College of Computer Science and Technology, Jilin University, Changchun, Jilin 100012, China, wucg@jlu.edu.cn.[4] Department Biomedical Informatics, College of Medicine, The Ohio State University, Columbus, OH 43210, USA, Qin.Ma@osumc.edu. [*] Corresponding author

**Abstract**

Support vector regression (SVR) is one of the most popular machine learning algorithms aiming to generate the optimal regression curve through maximizing the minimal margin of selected training samples, i.e., support vectors. Recent researchers reveal that maximizing the margin distribution of whole training dataset rather than the minimal margin of a few support vectors, is prone to achieve better generalization performance. However, the margin distribution support vector regression machines suffer difficulties resulted from solving a non-convex quadratic optimization, compared to the margin distribution strategy for support vector classification, This paper firstly proposes a maximal margin distribution model for SVR(MMD-SVR), then implementing coupled constrain factor to convert the non-convex quadratic optimization to a convex problem with linear constrains, which enhance the training feasibility and efficiency for SVR to derived from maximizing the margin distribution. The theoretical and empirical analysis illustrates the superiority of MMD-SVR. In addition, numerical experiments show that MMD-SVR could significantly improve the accuracy of prediction and generate more smooth regression curve with better generalization compared with the classic SVR.

## Introduction

Support vector machine (SVM) is one of the most popular learning algorithms for regression/classification, which maximize the minimal margin of training samples to generate an optimal regression/classification boundary, giving rise to the excellent generalization performance (Drucker and Burges 1997; Vapnik 1995). Furthermore, optimizing the margin of SVM/SVR will lead to quadratic programming (QP) following many efficient optimization algorithms (Drucker and Burges 1997). Inspired by the training criterion of SVM i.e. maximal margin, many margin-based algorithms are proposed to improve the generalization capability in recent years (Smola and Schölkopf 2004; Zhang and Zhou 2014; Shen et al. 2015; Xu et al. 2015; Liu et al. 2015; Lajugie et al. 2014; Kontorovich and Weiss 2014).

However, whether maximizing the minimal margin of training samples is the best way of exploring the margin theory. Studies on overfitting resistance of Adaboost (Freund and Schapire 1995) indicate well the answer to this problem. Specifically, Schapire et al. reported that AdaBoost has a surprising superiority to resist overfitting (Schapire et al. 1998); Breiman revealed that the essence for the anti-overfitting merit of Adaboost lies in maximizing the minimal margin (Breiman 1999). However, they fail to verify their assumption with the proper algorithm and experiment. Several years later, Reyzin et al. improved successfully Adaboost with margin distribution replacing the original minimal margin, getting the algorithm with better generalization capability (Reyzin and Schapire 2006). Recently, Gao and Zhou proved that maximizing the margin distribution, indicating margin mean and variance, is a more natural representation than only maximizing the minimal margin of training samples (Gao and Zhou 2013). Furthermore, Zhang and Zhou firstly applied margin distribution theory to SVM for classification, named large margin distribution machine (LMD) (Zhang and Zhou 2014), which endows classifier more generalization classification boundary than the canonical SVM.

As a main margin-based algorithm for regression, SVR absorbs extensive attention of researchers to enhance the generalization and noise resistance capabilities (Shao et al. 2013; Peng 2010; Yang et al. 2014; Balasundaram and Tanveer 2013; Balasundaram and Gupta 2014). However, the optimal predictor of SVR often are depended on a few training samples (i.e., support vectors), and it is hard that the smoothness of regression curve can be guaranteed by practical samples for data set with high noise and outlier. Intuitively, margin distribution could characterize the nature of feature space intensively and comprehensively. Motivated by the margin distribution theory (Gao and Zhou 2013; Zhang and Zhou 2014), this paper proposes the maximal margin distribution SVM for regression, (MMD-SVR), which introduces the mean and variance of margin for training set as margin distribution.

Actually, it should be pointed out that training MMD-SVR will give rise to a non-convex quadratic optimization problem with quadratic constraints. It would only adopt the linear kernel for linear regression and not be able to handle non-linear data set if semi-definite programming (SDP) is used to relax the training model. Here, we convert the original MMD-SVR model to a quadratic optimization problem by considering the geometric property of margin distribution in SVR, and then, linearizes all quadratic constraints by introducing relaxation with the coupled constrain factors; finally solves this QP model with linear constraints by an improved dual coordinate descent method (Hsieh et al. 2008). Numerical experiments in synthetic and practical benchmark data sets demonstrate that margin distribution is a much better form to characterize the sample margin in regression.

# Framework of Maximal Margin Distribution SVR

## Support Vector Regression Machine

We denote the training set

$$\boldsymbol{S} = \{(\boldsymbol{x}_1, y_1), (\boldsymbol{x}_1, y_1) \dots (\boldsymbol{x}_n, y_n)\}$$

where $\boldsymbol{x}_i \in R^d$ are training instances and $y_i \in R$ are the corresponding observation values. The target of regression is to learn a regression function, $f(\boldsymbol{x})$, to fit the real distribution of sample space, dependent on the training set $\boldsymbol{S}$, and predict the observation value for any new instance $\boldsymbol{x}$. SVR is a classic generalized linear regression method, which converts the non-linear regression in the function space to linear regression in the feature space by kernel mapping, i.e.

$$f(\boldsymbol{x}) = \boldsymbol{\omega} \cdot \phi(\boldsymbol{x}) + b = \sum_{i=1}^{l} (\alpha_i - \alpha_i^*) k(\boldsymbol{x}_i, \boldsymbol{x}) + b,$$

where $k(\boldsymbol{x}_i, \boldsymbol{x}_j) = \phi(\boldsymbol{x}_i) \cdot \phi(\boldsymbol{x}_j)$ represents the kernel mapping and $\alpha$ are dual factors. SVR gives rise to the optimal parameters by maximizing the minimal sample margin.

According to the literature (Drucker and Burges 1997; Vapnik 1995), the margin, $\tau_i$, of SVR for sample $(\boldsymbol{x}_i, y_i)$ is defined as

$$\tau_i = \frac{|f(x_i) - y_i|}{\|\omega\|} \le \frac{\varepsilon}{\|\omega\|} \quad (1)$$

where $\boldsymbol{\omega}$ is the vector of regression coefficients in the feature space, and $\varepsilon$ bound the margin tolerance.

Maximizing the minimal margin defined by support vectors which located on $\varepsilon$ boundary line for SVR and applying a soft margin to improve the generalization performance will result in the following optimization problem:

$$\begin{cases} \min \frac{1}{2} \|\boldsymbol{\omega}^2\| + C \sum_{i=1}^{n} (\xi_i + \xi_i^*) \\ s.t. \\ y_i - f(\boldsymbol{x}_i) \le \varepsilon + \xi_i \\ f(\boldsymbol{x}_i) - y_i \le \varepsilon + \xi_i^* \end{cases}, \quad (2)$$

where $\xi_i$, $\xi_i^*$ are the relaxation variables for sample $(\boldsymbol{x}_i, y_i)$ and represent to the residues of exceeding to the insensitive term $\varepsilon$ in the upper and lower boundary of $f(\boldsymbol{x}_i)$. Similar with SVM, SVR maximizes the $\varepsilon$ insensitive region (mean maximum margin) margin to generate the regression function $f(\boldsymbol{x}_i)$, which still is ordinary margin derived from support vectors, instead of the margin distribution of whole training set.

## Maximal Margin Distribution SVR

Here we consider the elementary statistics, the mean and the variance of margin, for representing margin distribution (Zhang and Zhou 2014), then:

$$\begin{cases} \tau_m = \frac{1}{n} \sum_{i=1}^{n} \tau_i = \frac{1}{n} \sum_{i=1}^{n} \frac{|f(x_i) - y_i|}{\|\omega\|} \\ \tau_v = \frac{1}{n} \sum_{i,j=1}^{n} (\tau_i - \tau_j)^2 = \frac{1}{n} \sum_{i,j=1}^{n} \frac{(|f(x_i) - y_i| - |f(x_i) - y_j|)^2}{\|\omega\|^2} \end{cases} \quad (3)$$

Taking the soft margin and $\varepsilon$ insensitive loss into consideration, according to (Gao and Zhou 2013; Zhang and Zhou 2014), maximizing the margin distribution is equal to maximizing margin mean $\tau_m$ and minimizing margin variance $\tau_v$ at the same time, which leads to the optimization problem (4):

$$\begin{cases} \min C \sum_{i=1}^{n} (\xi_i + \xi_i^*) - \lambda_1 * \tau_m + \lambda_2 * \tau_v \\ s.t. \\ y_i - f(\boldsymbol{x}_i) \le \varepsilon + \xi_i \\ f(\boldsymbol{x}_i) - y_i \le \varepsilon + \xi_i^* \end{cases}. \quad (4)$$

The optimization problem (4) is a non-convex quadratic programming problem (QPP) due to the regular term of $\tau_v$, and is hard to be solved directly. Generally, the approach to solve these problems is to get the corresponding convex relaxation formula making the solving feasible by some methods, such as SDP and quadratic cone programming, etc. Nevertheless, these methods exist obvious shortcomings, for example, the computational cost is too large, and kernel mapping is no longer available for non-linear problems.

Generally speaking, the illustration and the comparison canonical SVR with the proposed MMD-SVR could be given by Figure 1. Figure 1-a is the illustration of canonical SVR. The goal of canonical SVR is to cover all the data points concentrating in the belt region bounded by $\varepsilon$-non-sensitive functions, i.e., the region bounded by the two dashed lines. Figure 1-b is used to explain the MMD-SVR in an ideal situation, where the goal is to assemble all the data points lying exactly on the margin edges, i.e., the dashed lines, given by $\varepsilon$-non-sensitive functions. Figure 1-c is the interpretation the MMD-SVR in most practical situations, where the goal is to compel data points concentrating in two narrow belt regions bounded respectively by two dashed lines, marked with shadow, located at both sides of $f(x)$ and far away from $f(x)$ as much as possible within the $\varepsilon$ insensitive region. On each side, the narrow belt region is bounded by $\varepsilon$ (lines $L_1$ and $L_4$) and margin variance (lines $L_2$ and $L_3$). For the sake of mathematical convenience, the solid lines are denoted as $\mu\varepsilon$ ($0 \le \mu \le 1$), which are centered in both margin regions respective, corresponding to the margin mean.

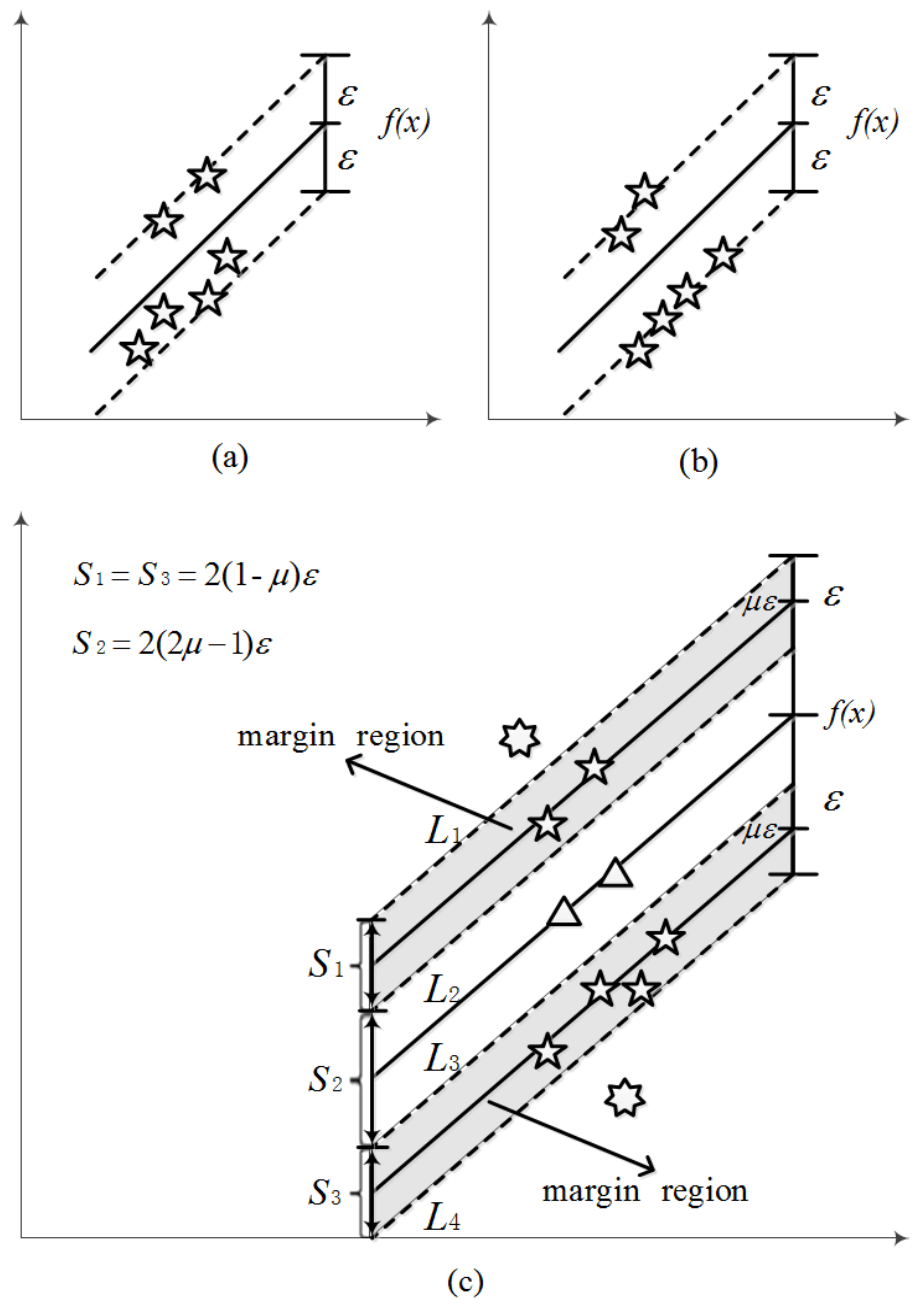

Figure 1: Illustration of MMD-SVR

# Maximal Margin Distribution SVR with Coupled Constraints

## Transformation of Maximal Margin Distribution SVR

**THEOREM 1**: Maximizing margin mean and minimizing margin variance within the $\varepsilon$ regions for MMD-SVR can be transformed to the following optimization problem (5).

$$\begin{cases} \min \frac{1}{2}\|\boldsymbol{\omega}\|^2 + C_1 \sum_{i=1}^{n}(\xi_i + \xi_i^*) + C_2 \sum_{i=1}^{n} \varsigma_i \\ s.t. \\ y_i - f(\boldsymbol{x}_i) \le \varepsilon + \xi_i \\ f(\boldsymbol{x}_i) - y_i \le \varepsilon + \xi_i^* \\ (f(\boldsymbol{x}_i) - y_i)^2 \ge (2\mu - 1)^2 * \varepsilon^2 - \varsigma_i \\ \xi_i \ge 0, \xi_i^* \ge 0, 0 \le \varsigma_i \le (2\mu - 1) * \varepsilon \end{cases} \quad (5)$$

where $\mu$ ($0 \le \mu \le 1$) is a constant, $(2\mu - 1)\varepsilon$ is the inner boundary of margin regions (i.e., $L_2$ and $L_3$ in figure 1-c), and $\varsigma_i$ are the penalty term for sample $(\boldsymbol{x}_i, y_i)$ which goes beyond the inner boundary of any margin region (S2 region in figure 1-c).

**Proof**: For any sample $(\boldsymbol{x}_i, y_i)$, its margin, $\tau_i$, is defined as:

$$\tau_i = \frac{|f(x_i) - y_i|}{\|\boldsymbol{\omega}\|}.$$

Due to Eq. (4), the mean and variance of margin for training set $S$ are:

$$\tau_m = \frac{1}{n}\sum_{i=1}^{n} \frac{|f(x_i) - y_i|}{\|\boldsymbol{\omega}\|} \le \frac{1}{n}\frac{n * \varepsilon}{\|\boldsymbol{\omega}\|} = \frac{\varepsilon}{\|\boldsymbol{\omega}\|},$$

$$\tau_v = \frac{1}{n}\sum_{i,j=1}^{n}(\tau_i - \tau_j)^2 = \frac{1}{n}\sum_{i,j=1}^{n} \frac{(|f(x_i) - y_i| - |f(x_j) - y_j|)^2}{\|\boldsymbol{\omega}\|^2} \ge 0.$$

In ideal situation, maximizing the margin mean and minimizing the margin variance will ideally lead to: $\tau_m = \frac{\varepsilon}{\|\boldsymbol{\omega}\|}$ and $\tau_v = 0$, which indicate that in ideal situation, $\tau_i = \frac{\varepsilon}{\|\boldsymbol{\omega}\|}$, as shown in figure 1-b, where all samples are distributed on the boundary lines determined by the $\varepsilon$ insensitive loss function instead of just support vectors in canonical SVR. Following the $L_2$ regularization of maximizing minimal margin for the canonical SVR, maximizing $\tau_m$ is equivalent to minimizing $\frac{1}{2}\|\boldsymbol{\omega}\|^2$, under the condition of all training samples located on the ε-boundaries.

However, as shown in figure 1-c, for most practical situations, due to the non-uniform sampling in sample space and noise existence, $\tau_m$ and $\tau_v$ only will approach, but not equal, to $\frac{\varepsilon}{\|\boldsymbol{\omega}\|}$ and 0, respectively.typically, for the specific $\mu\varepsilon$ in figure 1-c ,it can be derived: $\tau_m = \frac{\mu\varepsilon}{\|\omega\|}$ and $\tau_v = \left(\frac{(1-\mu)\varepsilon}{\|\omega\|}\right)^2$. Furthermore, we consider the Soft-margin of MMD-SVR to tolerate the outliers.

The margin mean gives rise to the following constraints:

$$\min \frac{1}{2}\|\boldsymbol{\omega}\|^2$$

and the soft constraints for margin variance give rise to the following constraints:

$$\begin{cases} y_i - f(\boldsymbol{x}_i) \le \varepsilon + \xi_i \\ f(\boldsymbol{x}_i) - y_i \le \varepsilon + \xi_i^* \\ \xi_i \ge 0, \xi_i^* \ge 0 \\ (f(\boldsymbol{x}_i) - y_i)^2 \ge (2\mu - 1)^2 * \varepsilon^2 - \varsigma_i \\ 0 \le \varsigma_i \le (2\mu - 1) * \varepsilon \end{cases}.$$

Hence, considering the objective function and constraints together, we have the optimization problem (5). THEOREM 1 is proved. ∎

Obviously, the optimization of practical maximal margin distribution will form two “suitable” margin regions located inside of the two $\varepsilon$ boundary lines and on both sides of regression function $f(x)$. The canonical SVR optimizes the margin by adjusting the position of ε boundaries to make sure most of the samples located within the region bounded the ε insensitive boundaries, which leads to a convex quadratic optimization within the closed set as shown in figure 1-a. However, the proposed MMD-SVR tunes the margin mean and variance of training samples to guarantee most of the samples located in two margin regions as shown shadow belts (S1 and S3) in Figure 1-c. Explicitly, MMD-SVR will derive a non-convex optimization due to the coupled margin regions, which is extremely hard to solve in the area of not infeasible. The rest part of this paper will be devoted to addressing how to conquer the non-convex optimization of MMD-SVR.

## Relaxation MMD-SVR with Coupled Constraint Variables

For the non-convex optimization problem (5), it is feasible to consider the MMD-SVR as two coupled constraint SVRs corresponding to the two margin regions. Hence, some coupled constraint variables are introduced to convert the non-convex optimization problem (5) to a convex quadratic optimization problem with linear constraints. The converted optimization problem is as follows:

$$\begin{cases} \min \frac{1}{2}\|\boldsymbol{\omega}\|^2 + C_1 \sum_{i=1}^{n}(\xi_i + \delta_i^*) + C_2 \sum_{i=1}^{n}(\delta_i * \xi_i^*) \\ s.t. \\ L_1: \quad y_i - (f(\boldsymbol{x}_i) + \mu\varepsilon) \le (1-\mu)\varepsilon + \xi_i \\ L_2: \quad (f(\boldsymbol{x}_i) + \mu\varepsilon) - y_i \le (1-\mu)\varepsilon + \xi_i^* \\ L_3: \quad y_i - (f(x_i) - \mu\varepsilon) \le (1-\mu)\varepsilon + \delta_i \\ L_4: \quad (f(\boldsymbol{x}_i) - \mu\varepsilon) - y_i \le (1-\mu)\varepsilon + \delta_i^* \\ \xi_i^* + \delta_i \ge 2(2\mu - 1)\varepsilon \\ \xi_i, \xi_i^* \ge 0, \delta_i, \delta_i^* \ge 0 \end{cases} \tag{6}$$

where $L_1$ and $L_4$ restrict the samples of training set to the interior of closed region bounded by ε boundaries arising from ε-insensitive loss function, and the $\xi_i$, $\delta_i^*$ are the penalty terms for $L_1$ and $L_4$ with the soft margin, which are similar with those relaxation factors in the canonical SVR; $L_2$ and $L_3$, named coupled constraints, which restrict samples to the exterior of the middle irrelevant region generated by margin variance boundaries with $(2\mu - 1)\varepsilon$ width (S2 region in Figure 1-c), those two constraints are coupled with each other for punishing samples in opposite margin regions, that is, one margin region considers the "good" samples in its opposite margin region as outliers. Constraint pairs, ($L_1$, $L_2$) and ($L_3$, $L_4$) generate respectively two symmetrical margin regions on both sides of $f(x)$, centered on the margin mean $\tau_m$ and bounded by ε boundaries and $\tau_v$ (as shown the shadow regions in figure 1-c). However, the penalty terms of $\xi_i^*$and $\delta_i$ may produce partial overlap in the feasible regions bounded by ($L_1$, $L_2$), and ($L_3$, $L_4$), respectively, i.e., the large $\xi_i^*$ perhaps allow samples to locate in the margin region S3 bounded by $L_3$ and $L_4$, due to the existence of coupled constraints $L_2$ and $L_3$. Samples fall into this kind of overlap region, i.e., S1 for $\delta_i$, S3 for $\xi_i^*$ in Figure 1-c should not be punished. Hence, we introduce the coupled constraint factors, i.e., $\delta_i * \xi_i^*$, to eliminate the influence of these undesirable penalties and, meanwhile, to make sure that outliers located in the region bounded by $L_2$ and $L_3$are punished. Notice that $\xi_i^* + \delta_i$ is equal to $2(2\mu - 1)\varepsilon$ when the sample $(\boldsymbol{x}_i, y_i)$ locates in the ($L_2$, $L_3$)region; otherwise, $\xi_i^* + \delta_i$ should be larger than $2(2\mu - 1)\varepsilon$. Hence, we impose the fifth constraint, $\xi_i^* + \delta_i \ge 2(2\mu - 1)\varepsilon$.

**THEOREM 2**: The term of coupled constraint factors $\sum_{i=1}^{n}(\delta_i * \xi_i^*)$ gives rise to the minimal value when the optimization problem (5) is optimal.

**Proof**: The training samples can be classified into two classes depend on their relative positions in the feature space: the first class contains those samples locating in the margin region or beyond the $\varepsilon$ insensitive line. For those samples of the first class, generally, assuming the sample $(\boldsymbol{x}_i, y_i)$ is located in the ($L_1$, $L_2$) region, according to the KKT condition, $\xi_i^* = 0$ and $\delta_i \ge 0$ hold, when the optimization problem (6) is optimal; otherwise, if it is located in the ($L_3$, $L_4$) region, optimal solutions of the optimization problem (6) make both $\xi_i^* \ge 0$ and $\delta_i = 0$ hold. Hence, the coupled constraint factors $\delta_i * \xi_i^* = 0$ hold naturally for samples of the first class.

The second class contains those samples locating in the ($L_2$, $L_3$) region, for example, the green stars in Figure 1-c. Notice that the width of the penalty region ($L_2$, $L_3$) is equal to $2(2\mu - 1)\varepsilon$, i.e., S2 shown in Figure 1-c. For those samples of the second class, according to the KKT condition, when the optimization problem (6) is optimal, we have

$$\xi_i^* \ge 0, \delta_i \ge 0, \text{ and } \xi_i^* + \delta_i = 2(2\mu - 1)\varepsilon.$$

Then, it turns out

$$0 \le \delta_i * \xi_i^* \le \frac{(\delta_i + \xi_i^*)^2}{2} = 2(2\mu - 1)^2\varepsilon^2.$$

The coupled constraint factor $\delta_i * \xi_i^*$ acquires maximum if and only if $\delta_i = \xi_i^* = (2\mu - 1)\varepsilon$. And notice that $\xi_i^*$ can be represented as $\xi_i^* = 2(2\mu - 1)\varepsilon - \delta_i$, then

$$\delta_i * \xi_i^* = \delta_i * (2(2\mu - 1)\varepsilon - \delta_i)$$

$$\delta_i * \xi_i^* = \delta_i * (2(2\mu - 1)\varepsilon - \delta_i) = -(\delta_i - (2\mu - 1)\varepsilon)^2 + (2\mu - 1)^2\varepsilon^2.$$

It is easy to know that when $\delta_i$ increases from 0 to $(2\mu - 1)\varepsilon$, $\delta_i * \xi_i^*$ increases monotonously from 0 to $(2\mu - 1)^2\varepsilon^2$. Then we have

$$\delta_i * \xi_i^* = k_i \varsigma_i$$

Following the KKT condition, we can get the equivalent representation for quadratic constraint term of formula 5:

$$\varsigma_i = (2\mu - 1)^2 * \varepsilon^2 - (f(\boldsymbol{x}_i) - y_i)^2$$
$$= (2\mu - 1)^2 * \varepsilon^2 - \left((2\mu - 1)\varepsilon - r_i\right)^2$$

Where $r_i$ is the penalty term for sample $(\boldsymbol{x}_i, y_i)$ in ($L_2$, $L_3$) region, so $r_i = min(\delta_i, \xi_i^*)$, and assume $min(\delta_i, \xi_i^*) = \delta_i$, then

$$\delta_i * \xi_i^* = \varsigma_i$$

Jointing with the situations of both the first and the second classes, THEOREM 2 is proved.

## Model Solving

The Lagrangian of the minimal optimization problem (6) takes the following form:

$$
\begin{aligned}
L(\omega,\xi,\xi^*,\delta,\delta^*,\theta) = {} & \frac{1}{2}\omega^T\omega + C_1\sum_{i=1}^n(\xi_i+\delta_i^*) \\
& +C_2\sum_{i=1}^n(\delta_i * \xi_i^*) \\
& -\sum_{i=1}^n \alpha_i\big((1-\mu)\varepsilon + \xi_i - y_i + (f(\boldsymbol{x}_i)+\mu\varepsilon)\big) \\
& -\sum_{i=1}^n \alpha_i^*\big((1-\mu)\varepsilon + \xi_i^* - (f(\boldsymbol{x}_i)+\mu\varepsilon) + y_i\big) \\
& -\sum_{i=1}^n \beta_i\big((1-\mu)\varepsilon + \delta_i - y_i + (f(\boldsymbol{x}_i)-\mu\varepsilon)\big) \\
& -\sum_{i=1}^n \beta_i^*\big((1-\mu)\varepsilon + \delta_i^* - (f(\boldsymbol{x}_i)-\mu\varepsilon) + y_i\big) \\
& +\sum_{i=1}^n \psi_i(\xi_i^* + \delta_i + 2*(2\mu-1)\varepsilon) \\
& -\sum_{i=1}^n(\lambda_i\xi_i + \lambda_i^*\xi_i^* + \eta_i\delta_i + \eta_i^*\delta_i^*)
\end{aligned} \quad (7)
$$

By setting the partial derivations of $\omega,\xi,\xi^*,\delta,\delta^*,\theta$ to zero, primal variables $\omega,\xi,\xi^*,\delta,\delta^*,\theta$ can be represented by dual factors, such as $\alpha,\alpha^*,\beta,\beta^*,\psi$. By substituting the dual form of $\alpha,\alpha^*,\beta,\beta^*,\psi$ to the Lagrangian $L(\omega,\xi,\xi^*,\delta,\delta^*)$, we get the dual optimization problem of the primal problem (6) as followings:

$$
\begin{cases}
\min_{\boldsymbol{\alpha},\boldsymbol{\alpha}^*,\boldsymbol{\beta},\boldsymbol{\beta}^*} \frac{1}{2}(\boldsymbol{\alpha}-\boldsymbol{\alpha}^*+\boldsymbol{\beta}-\boldsymbol{\beta}^*)^T\phi(\boldsymbol{X})^T\phi(\boldsymbol{X})(\boldsymbol{\alpha}-\boldsymbol{\alpha}^*+\boldsymbol{\beta}-\boldsymbol{\beta}^*) \\
\quad -(\boldsymbol{\alpha}-\boldsymbol{\alpha}^*+\boldsymbol{\beta}-\boldsymbol{\beta}^*)^T\boldsymbol{Y} + \frac{1}{C_2}(\boldsymbol{\alpha}^*+\boldsymbol{\psi})^T(\boldsymbol{\beta}+\boldsymbol{\psi}) \\
\quad +(\boldsymbol{\alpha}^*+\boldsymbol{\beta}+2\boldsymbol{\psi})^T(1-2\mu)\varepsilon\boldsymbol{I} \qquad (8) \\
s.t. \quad 0 \le \boldsymbol{\alpha},\boldsymbol{\beta}^* \le C_1 \\
\quad 0 \le \boldsymbol{\alpha}^*+\boldsymbol{\psi}, \boldsymbol{\beta}+\boldsymbol{\psi} \le C_2 \\
\quad \boldsymbol{\alpha}^* \ge 0, \boldsymbol{\beta} \ge 0, \boldsymbol{\psi} \ge 0
\end{cases}
$$

The problem (8) is quadratic optimization with linear and box constraints. According to (Hsieh et al. 2008), generally speaking, the problem (8) could be solved by dual coordinate descent method. Hence, we only need to develop an improved method to deal with coupled constraint variables.

Specifically, one of the variables is selected to minimize while the other variables are kept as constants at each iteration. Assume the current optimization variable is $\alpha_i$, we have

$$
\begin{cases}
f(\alpha_i+t) = \frac{1}{2}\phi(\boldsymbol{x}_i)^T\phi(\boldsymbol{x}_i)t^2 + \frac{\partial f(\alpha_i)}{\partial \alpha_i}t + f(\alpha_i) \\
s.t. \quad 0 \le \alpha_i \le C_1
\end{cases} \quad (9)
$$

As the derivations $\frac{\partial f(\alpha_i)}{\partial \alpha_i}$ independent on $t$, then Eq. (9) can be considered as a quadratic function of $t$. Based on the theory of optimization, $\min_t f(\alpha_i+t)$ will give rise to the optimal solution in conditions of $\frac{\mathrm{d}f(\alpha_i+t)}{\mathrm{d}t}=0$ and $\phi(\boldsymbol{x}_i)\cdot\phi(\boldsymbol{x}_i)=k(x_i,x_i)$. Then we have:

$$t_{opt} = \frac{\frac{\partial f(\alpha_i)}{\partial \alpha_i}}{k(x_i,x_i)}.$$

Due to the box constraint $0 \le \alpha \le C_1$, the minimizer of Eq. (9) leads to a closed-form solution:

$$\alpha_i^{new} = min(max(\alpha_i - t_{opt}, 0), C_1). \quad (10)$$

The optimization of variable $\boldsymbol{\beta}^*$ is similar to $\boldsymbol{\alpha}$. Furthermore, the optimization for the variable $\boldsymbol{\alpha}^*$and $\boldsymbol{\beta}$ is a little different with $\boldsymbol{\alpha}$ for upbound of box constraint dependent on the auxiliary $\psi$.

$$0 \le \beta_i^{new} \le C_2 - \psi.$$

To make sure that $\beta_i^{new}$ is feasible, we have:

$$\beta_i^{new} = \sup_k(\beta_i - t_{opt} * \upsilon^k) \le C_2 - \psi, \quad (11)$$

where $0 < \upsilon \le 1, k \in N$.

The auxiliary variable, $\psi$, can be solved in the feasible region by Newton's method, i.e.,

$$f(\psi_i+t) = \frac{1}{C_2}t^2 + \frac{\partial f(\psi_i)}{\partial \psi_i}t + f(\psi_i),$$

$$t_{opt} = \frac{C_2\frac{\partial f(\alpha_i)}{\partial \alpha_i}}{2},$$

$$\psi_i^{new} = \sup_k(\psi_i - t_{opt} * \upsilon^k \le C_2 - \max(\alpha^*,\beta), \quad (12)$$

where $0 < \upsilon \le 1, k \in N$.

# Experiments

## Experimental Settings

In this section, we empirically check the performance of MMD-SVR on several data sets, including six synthetic data sets and nine practical data sets. Both experiments on synthetic data sets and practical data sets are implemented with 5-fold cross-validation and repeated 10 times independently. Each attribute of data sets is normalized into the interval [-1, 1]. To compare the performance of our proposed MMD-SVR, the coefficient of determination denoted as $R^2$, is used as a comparison criterion, i.e., which is defined as follows (Kvalseth 1985):

$$R^2 = 1 - \left(\frac{med(|y_i - f(x_i)|)}{mad(y_i)}\right)^2, mad(y_i) = med(|y_i - med(y_i)|).$$

We compared MMD-SVR with canonical SVR (Drucker et al. 1997) and ε-twin support vector machine for regression (ETSVR) (Shao et al. 2013), gauss process of machine learning (GPML) (Blake and Merz 1998) and robust least squares support vector machine for regression (RLS-SVM) (Yang et al. 2014). All the parameters of MMD-SVR, canonical SVR, ETSVR, GPML, and RLS-SVM are selected by 5-fold cross validation from their finite parameter sets. For RLS-SVM and canonical SVR, the parameter $C$ is selected from $\{2^0, 2^1, 2^2, \cdots, 2^9\}$ (Balasundaram and Gupta 2014). For ETSVR, the parameters $c_1, c_2, c_3, c_4, \varepsilon_1$ and $\varepsilon_2$ are selected from the set of $\{2^{-8}, 2^{-7}, 2^{-6}, \cdots, 2^8\}$. For MMD-SVR, the parameters $\mu$, $C_1$ and $C_2$ are selected $\{0.5, 0.6, 0.7, \cdots, 1\}$ for $\mu$and $\{5, 10, 15, \cdots, 30\}$ for last two, respectively. For MMD-SVR, canonical SVR, ETSVR, GPML and RLS-SVM, the width of the RBF kernel is selected $\{2^{-4}\delta, 2^{-3}\delta, 2^{-2}\delta, \cdots, 2^5\delta\}$, where δ is the average distance between the instances. For GPML, the optimal parameters of mean and covariance functions are selected by the conjugate gradients algorithm with 100 iterations.

## Comparison results on Synthetic Dataset

There are six benchmark functions which are selected from (Peng 2010; Balasundaram and Tanveer 2013; Balasundaram and Gupta 2014). The comparison results are listed in Table 1. 200 samples are generated randomly according to their definition domains, and training samples are mixed with Gaussian noise with 0 mean and $0.1^2$ variance. From Table 1, we can observe that MMD-SVR gets the biggest $R^2$ on three testing problems and obtains the biggest average $R^2$ for all testing problems. The numbers of win cases of MMD-SVR, with respect to SVR, ETSVR, GPML, RLS-SVM, are 6, 5, 3 and 2, respectively. The win/tie/loss counts show also that MMD-SVR is very competitive to all compared methods.

## Comparison results on Real Benchmark Datasets

Here, data sets Auto MPG, Concrete CS, Yacht hydrodynamics, Boston housing, and Wine quality red are UCI data sets (Blake and Merz 1998), the data sets Chwirut1, Nelson, Baby fat and Gauss3 are downloaded from http://www.itl.nist.gov/div898/strd/nls/main.shtml. All the data sets are commonly used in test machine learning algorithms. Moreover, the comparison results are listed in Table 2. Obviously, MMD-SVR outperforms all compared methods. MMD-SVR gets the biggest $R^2$ on 6 datasets and obtains the biggest average $R^2$. Moreover, MMD-SVR performs significantly better than SVR, ETSVR, GPML, RLS-SVM on data sets with the number of 7, 6, 6 and 7, respectively. In addition, the win/tie/loss counts show that MMD-SVR is very competitive.

Here, the results corresponding to the biggest $R^2$ of each benchmark function is bolded. •/○ indicates that the performance is significantly better/worse than SVR (paired t-test at a 95% significance level). The win/tie/loss counts for MMD-SVR are summarized in the last rows.

Table 1: Comparison on benchmark functions

| Data set | SVR | ETSVR | GPML | RLS-SVM | MMD-SVR |
|---|---|---|---|---|---|
| Function1 | 0.7758±0.1148 | 0.9997±6.4793e-04 • | 0.9999±7.4492e-05 • | 1.0000±4.5193e-08 • | **1.0000±3.9799e-08** • |
| Function2 | 0.9240±0.0270 | 0.9986±7.4793e-04 • | 0.9994±4.5716e-04 • | 1.0000±3.0487e-06 • | **1.0000±2.1939e-06** • |
| Function3 | 0.8340±0.0971 | 0.9961±0.0028 • | 0.9999±8.0106e-05 • | **1.0000±1.6571e-07** • | 1.0000±6.1949e-06 • |
| Function4 | 0.9373±0.0329 | 0.9814±0.0175 • | 0.9436±0.1259 | **1.0000±2.0433e-05** • | 0.9999±4.5502e-05 • |
| Function5 | 0.2205±0.2087 | 0.9890±0.0033 • | -0.1467±0.3370 ○ | 0.9955±0.0010 • | **0.9988±5.7786e-04** • |
| Function6 | 0.9735±0.0107 | 0.9966±0.0020 • | **1.0000±1.0563e-09** • | 0.9993±3.6424e-04 • | 0.9990±6.9105e-04 • |
| Ave. $R^2$ | 0.7775 | 0.9936 | 0.7994 | 0.9991 | 0.9996 |
| win/tie/loss | 6/0/0 | 5/1/0 | 3/2/1 | 2/3/1 | |

Table 2: Comparison on benchmark data sets

| Data set | SVR | ETSVR | GPML | RLS-SVM | MMD-SVR |
|---|---|---|---|---|---|
| Auto MPG | 0.9494±6.4415e-04 | 0.9461±2.4404e-04 ○ | 0.9276±0.0012 ○ | 0.8580±0 ○ | **0.9595±7.9565e-04** • |
| Concrete CS | 0.8793±0.0220 | 0.9073±0.0229 | 0.8847±0.0252 | 0.9145±0.0269 | **0.9431±0.0086** • |
| Yacht H.D. | 0.4186±0.1133 | 0.9279±0.0294 • | **0.9988±6.5226e-04** • | 0.8154±0.0689 • | 0.9963±9.2792e-04 • |
| Bodyfat | 0.9671±0.0061 | 0.9944±0.0017 • | **0.9996±1.5680e-04** • | 0.6863±0.0646 ○ | 0.9882±0.0038 • |
| Wine QE | 0.8764±0.0130 | **0.9027±0.0123** • | 0.8487±0.0225 ○ | 0.6318±0.1433 ○ | 0.8976±0.0188 |
| Chwirut1 | 0.9359±0.0025 | 0.9582±5.7596e-04 • | 0.9608±1.2413e-16 • | 0.9790±1.2413e-16 • | **0.9810±0.0011** • |
| Gauss3 | 0.9917±3.6714e-04 | 0.9928±1.6800e-04 • | 0.9938±0.0033 | 0.9973±1.2413e-16 • | **0.9984±2.1056e-04** • |
| Boston H. | 0.9418±0.0016 | 0.9435±7.3863e-04 | 0.9227±0.0036 ○ | 0.8298±1.2413e-16 ○ | **0.9529±9.0945e-04** • |
| Nelson | 0.6546±0.2502 | 0.6903±0.2778 | 0.7247±0.2439 | 0.7433±0.1346 | **0.8188±0.1972** |
| Ave. $R^2$ | 0.8461 | 0.9181 | 0.9179 | 0.8284 | 0.9484 |
| win/tie/loss | 7/2/0 | 6/2/1 | 6/1/2 | 7/2/0 | |

# Conclusion

In this paper, we work on SVM for regression problems by maximizing the margin distribution and propose maximal margin distribution support vector regression (MMD-SVR), which maximizes the margin distribution with the whole training set, and then we propose the convex relaxation solver for the non-convex QCQP generated by MMD-SVR.

## References

Drucker, H; Burges, C. J. C.; Kaufman, L, et al. 1997. Support vector regression machines. *Advances in neural information processing systems*, 155-161.

Smola, A. J.; Schölkopf, B. 2004. A tutorial on support vector regression. *Statistics and computing* 14(3):199-222.

Zhang, T.; Zhou, Z. H. 2014. Large margin distribution machine. *In Proceedings of the 20th ACM SIGKDD international conference on Knowledge discovery and data mining*. 313-322.ACM

Shen, B., et al. 2015. SP-SVM: Large margin classifier for data on multiple manifolds. *Twenty-Ninth AAAI Conference on Artificial Intelligence (AAAI).*

Xu, C.; Tao, D; Xu, C. 2015. Large-Margin Multi-Label Causal Feature Learning. *Twenty-Ninth AAAI Conference on Artificial Intelligence (AAAI)*.

Liu, W.; Ivor W. 2015. Tsang. Large Margin Metric Learning for Multi-label Prediction. *Twenty-Ninth AAAI Conference on Artificial Intelligence (AAAI)*.

Lajugie, R.; Bach, F.; Arlot, S. 2014. Large-margin metric learning for constrained partitioning problems. *In Proceedings of The 31st International Conference on Machine Learning (ICML-14)*, 297-305.

Kontorovich, A.; Weiss, R. 2014. Maximum Margin Multiclass Nearest Neighbors. *arXiv preprint arXiv:1401.7898*.

Freund, Y.; Schapire, R. E. 1995. A decision-theoretic generalization of on-line learning and an application to boosting. *In Proceedings of the 2nd European Conference on Computational Learning Theory*, 23–37.

Schapire, R. E.; Freund, Y.; Bartlett, P. L.; Lee, W. S. 1998. Boosting the margin: a new explanation for the effectives of voting methods. *Annuals of Statistics* 26(5):1651–1686.

Breiman, L. 1999. Prediction games and arcing classifiers. *Neural Computation* 11(7):1493–1517.

Reyzin, L.; Schapire, R. E. 2006. How boosting the margin can also boost classifier complexity. *In Proceedings of 23rd International Conference on Machine Learning*, 753–760.

Gao, W.; Zhou, Z. H. 2013. On the doubt about margin explanation of boosting. *Artificial Intelligence* 203(10):1-18.

Vapnik, V. 1995. The Nature of Statistical Learning Theory. Mass.: *Springer-Verlag*.

Hsieh, C. J.; Chang, K. W.; Lin, C. J.; Keerthi, S. S.; Sundararajan, S. 2008. A dual coordinate descent method for large-scale linear svm. *In Proceedings of the 25th International Conference on Machine Learning*, 408–415.

Luo, Z. Q.; Tseng, P. 1992. On the convergence of coordinate descent method for convex differentiable minimization. *Journal of Optimization Theory and Application* 72(1):7–35.

Shao, Y. H.; Zhang, C. H.; Yang, Z. M.; Jing, L.; Deng, N. Y. 2013. An epsilon-twin support vector machine for regression. *Neural Computing and Applications* 23(1):175–185.

Peng, X. 2010. TSVR: an efficient twin support vector machine for regression. *Neural Networks* 23(3):365-372.

Yang, X.; Tan, L.; He. L. 2014. A robust least squares support vector machine for regression and classification with noise. *Neurocomputing* 140(9):41-52.

Rasmussen, C. E.; Williams, C. K. I. 2006. Gaussian processes for machine learning. *The MIT Press*.

Kvalseth, T.O. 1985. Cautionary Note about $R^2$. *The American Statistician* 39(4):279–285.

Balasundaram, S; Tanveer, M. 2013. On Lagrangian twin support vector regression. *Neural Computing & Applications* 22(5):257-267.

Balasundaram, S.; Gupta, D. 2014. Training Lagrangian twin support vector regression via unconstrained convex minimization. *Knowledge-Based Systems* 59(3):85-96.

Blake, C. L.; Merz, C. J. 1998. UCI repository for machine learning databases. *Department of Information and Computer Sciences, University of California, Irvine*.